\PassOptionsToPackage{capitalize}{cleveref}
\documentclass[runningheads]{llncs}
\PassOptionsToPackage{table,xcdraw}{xcolor}
 
\usepackage{eccv}

\usepackage{eccvabbrv}
\usepackage{algorithm}
\usepackage{algpseudocode}

\usepackage{graphicx}
\usepackage{booktabs}
\usepackage{xcolor}
\usepackage{soul} 

\colorlet{myyellow}{yellow!40}
\colorlet{mygreen}{green!25}
\newcommand{\hly}[1]{\sethlcolor{myyellow}\hl{#1}}
\newcommand{\hlg}[1]{\sethlcolor{mygreen}\hl{#1}}

\usepackage{multirow}

\usepackage{makecell}

\definecolor{bestgray}{gray}{0.85}
\newcommand{\up}[1]{{\scriptsize\textcolor[rgb]{0,0.5,0.3}{$\uparrow$#1}}}

\usepackage[accsupp]{axessibility}  

\usepackage[pagebackref,breaklinks,colorlinks,citecolor=eccvblue]{hyperref}
\usepackage{cleveref}
\usepackage{orcidlink}

\begin{document}

\title{SkillGraph: Self-Evolving Multi-Agent Collaboration with Multimodal Graph Topology} 

\titlerunning{SkillGraph}

\author{Zheng Nie\inst{1} \and
Ruolin Shen\inst{2} \and
Xinlei Yu\inst{1} \and
Bo Yin\inst{1} \and
Jiangning Zhang\inst{3} \and
Xiaobin Hu\inst{1}
}

\authorrunning{Z.~Nie et al.}

\institute{National University of Singapore, Singapore \and
Technical University of Munich, Munich, Germany \and
Zhejiang University, China}

\maketitle

\begin{abstract}
Scaling vision-language models into Visual Multiagent Systems (VMAS) is hindered by two coupled issues. First, communication topologies are fixed before inference, leaving them blind to visual content and query context; second, agent reasoning abilities remain static during deployment. These issues reinforce each other: a rigid topology fails to leverage richer agent expertise, while static agents lack incentives to specialize for a given query. We address this with SkillGraph, a joint framework that evolves both agent expertise and communication topology. Within this framework, a Multimodal Graph Transformer (MMGT) encodes visual tokens, instruction semantics and active skill embeddings to predict a query-conditioned collaboration graph, replacing hand-crafted routing with dynamic, content-aware information flow. Complementing this, a Skill Designer distills and refines reasoning heuristics from failure cases, constructing a self-evolving multimodal Skill Bank. Crucially, updated skill embeddings are fed back into the MMGT, enabling the topology to adapt alongside capability growth. Experiments show that SkillGraph achieves consistent improvements across four benchmarks, five common MAS structures and four base models. Code is available at https://github.com/niez233/skillgraph.

\end{abstract}

\section{Introduction}
\label{sec:intro}

\begin{figure}[t]
  \centering
  \includegraphics[width=\linewidth]{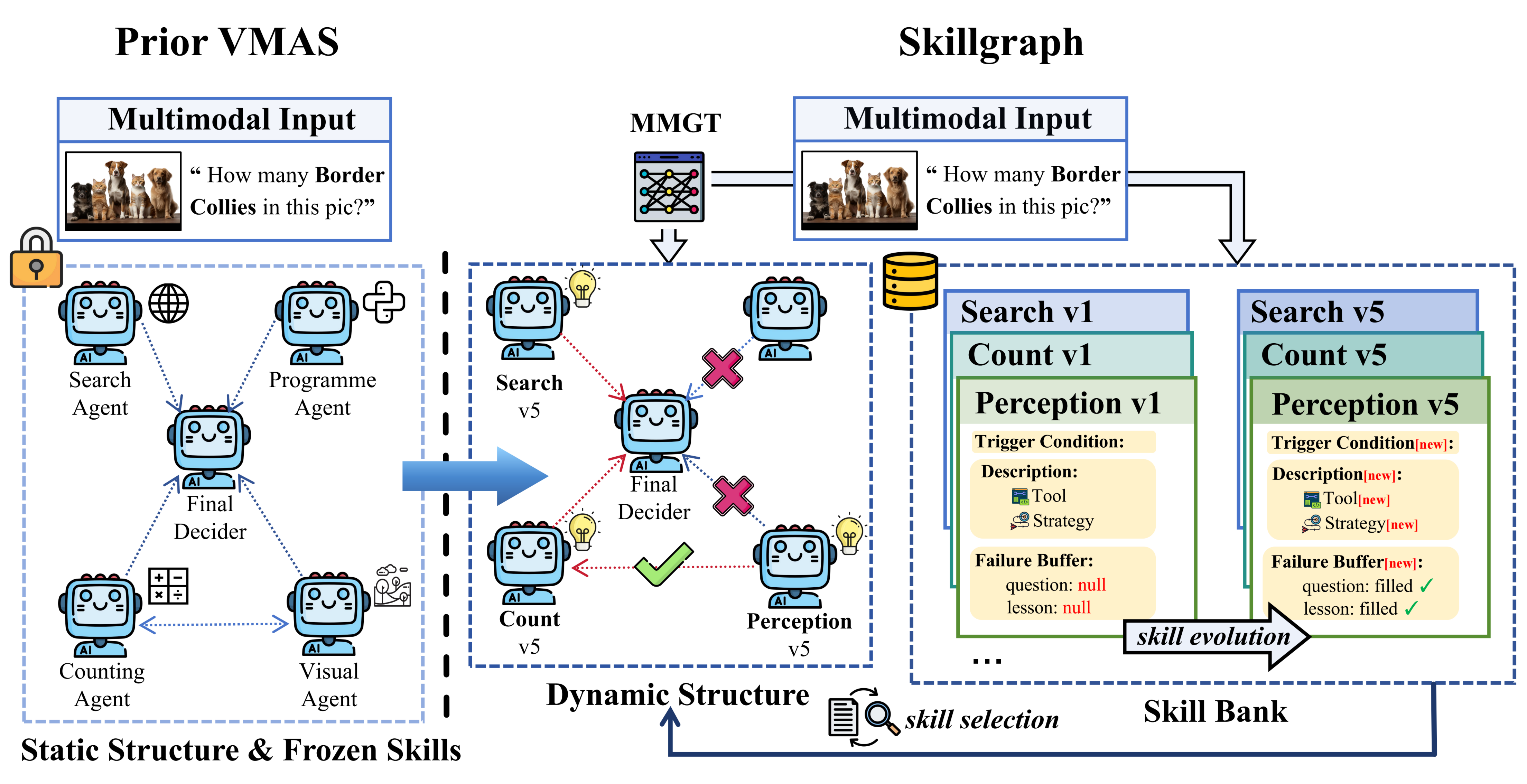}
  \caption{\textbf{Comparison of VMAS paradigms.} Prior VMAS uses static topologies and frozen skills. Our SkillGraph enables a co-evolution loop: MMGT predicts dynamic collaboration graphs, while a skill bank self-evolves agent capabilities.}
  \label{fig:intro}
\end{figure}

The rapid development of vision-language models (VLMs) has advanced single-model perceptual and reasoning capabilities. Consequently, research is shifting from a single-agent paradigm to Visual Multi-Agent Systems (VMAS) to leverage collective intelligence \cite{li2023camel, yu2025visual,hong2023metagpt, qian2024scaling, wu2024autogen,yu2026dual}. The core hypothesis is that VMAS, by wiring together specialized agents into a collaborative network, can yield substantial performance gains on complex, multi-step multimodal tasks that remain intractable for individual models. In an ideal VMAS framework, agents form a dynamic ensemble of experts whose communication structure is tailored to the multimodal characteristics of each query, enabling more effective reasoning trajectories.

However, as we scale these visual multi-agent collectives, a fundamental bottleneck emerges: the structural and cognitive rigidity of current VMAS frameworks. Existing systems primarily rely on static, predefined topologies and fixed role-playing pipelines \cite{zhang2024g, zhuge2024gptswarm}. These communication structures are established before the specific multimodal content of a query is even analyzed. Consequently, whether a task requires fine-grained OCR, complex spatial reasoning, or abstract logical deduction, the system often routes information through the same invariant pathways. This lack of adaptability is twofold. First, the topology is fixed a priori; instead of being jointly determined by the query’s multimodal semantics and the agents’ active skills, it routes information through invariant pathways that may not suit the specific task. Second, the expertise of the agents suffers from skill stagnation; their reasoning capabilities are frozen in hand-crafted text prompts. When the system encounters persistent multimodal "blind spots" at test time, there is no mechanism for agents to autonomously refine their skills or generate new reasoning heuristics.

We attribute this failure to the systemic decoupling of task content, agent skill capabilities and communication topology. Theoretically, the optimal collaboration graph for a visual task should not be a static template. Instead, it must be a dynamic function of both the query's unique multimodal semantics and the agents' active skills. The "wiring" of a VMAS should be determined only after the system understands what the specific task demands and which skills are currently most effective for those demands. However, in current architectures, the graph designer lacks the multimodal perception to "see" the task content, and the agents lack the plasticity to evolve their skills. For VMAS to overcome this bottleneck, the framework must bridge this gap by enabling a content-aware and skill-driven orchestration: synthesizing an optimized communication topology conditioned jointly on the query (image and text) and the active skill bank, while allowing the skills themselves to improve through interaction experience.

To address these challenges, we propose SkillGraph, a unified framework that enables the simultaneous evolution of agent expertise and communication topology in a closed loop. Our framework introduces the Multimodal Graph Transformer (MMGT), which replaces hand-crafted wiring with a learned, content-aware designer. MMGT jointly encodes visual tokens and instruction semantics for each query to predict a tailored communication topology, ensuring that information routing is directly aligned with the task requirements and the agents’ active skill representations. In parallel, we introduce a Self-Evolving Skill Bank for multimodal agents, where a Skill Designer module continuously refines reasoning heuristics from accumulated failure cases. Crucially, SkillGraph closes the co-evolution loop: as agent skills evolve and become more specialized, their updated representations propagate directly back into the MMGT. This allows the topology predictor to dynamically adapt its routing strategy to the agents' enhanced skill capabilities, making the structure and knowledge of the VMAS mutually reinforcing.

We validate SkillGraph across diverse VLM backbones and benchmarks, demonstrating its consistent superiority over fixed-topology and static-skill baselines. Our contributions are summarized as follows:
\begin{itemize}
\item \textbf{Skill-Conditioned Agents:} We construct a multimodal agent network by equipping each agent with a dynamically retrieved skill from a hierarchical Skill Bank and encoding the active skill into node features to reflect the agent's current reasoning state.
\item \textbf{MMGT Topology Design:} We introduce Multimodal Graph Transformer (MMGT), which jointly models visual tokens, question semantics, and role priors to predict a query-conditioned, directed topology.
\item \textbf{Self-Evolving Skill Bank:} We propose a Skill Designer that diagnoses recurring failures and uses them to modify or create skills, and feeds updated skill representations back into MMGT, forming a closed-loop co-evolution between agent capability and communication structure.
\end{itemize}

\section{Related Work}

\subsection{Multi-Agent Systems as Graphs}
Organizing LLM agents into structured collaboration graphs has progressed from static, role-fixed pipelines~\cite{li2023camel,
qian2024chatdev,wu2024autogen,holt2023l2mac,yu2025visual1} toward dynamically optimizable topologies. Early graph-based frameworks showed that encoding human workflows into DAG-structured agent networks consistently outperforms single-agent baselines~\cite{qian2024scaling,%
hong2023metagpt}, while debate-style topologies demonstrated the value of diverse agent perspectives~\cite{chan2023chateval,kim2024mdagents,du2024improving,liang2024encouraging,%
chen2024reconcile}. A pivotal shift came with jointly learnable topologies: GPTSwarm~\cite{zhuge2024gptswarm} used RL to co-optimize node prompts and edge connectivity; G-Designer~\cite{zhang2024g} introduced a variational graph
auto-encoder for query-adaptive topology prediction; and MASS~\cite{zhou2025multi} revealed that prompt and topology search are mutually reinforcing. Preceding these learnable methods, DyLAN~\cite{liu2023dynamic} and DSPy~\cite{khattab2023dspy} laid
important groundwork through dynamic team optimization and compiled agent pipelines, respectively. More recent work trains meta-agents to generate query-conditioned workflows end-to-end via RL~\cite{gao2025flowreasoner,dang2025multi,
hu2024automated}, or enables the graph itself to self-evolve through test-time feedback~\cite{hu2024self,xue2025comas}. Despite this progress, a critical gap persists: every existing method constructs the agent graph purely from text, leaving the visual content of the query outside the topology-prediction loop. SkillGraph closes this gap by conditioning the graph transformer jointly on question semantics and per-agent image attention, so that the inferred communication topology is a direct function of multimodal query content.

\subsection{Self-Improving Agents with Skill Libraries}
Recent work focuses on how agents accumulate, refine and reuse experience through explicit skill libraries rather than relying solely on gradient updates. Early approaches store reusable experience as natural-language memory summaries~\cite{zhao2024expel,wang2024agent}, executable code skills~\cite{wang2023voyager}, and abstracted workflow templates~\cite{zhang2024aflow}. Building on this line, recent skill-centric agent frameworks have shown that reusable skills can improve performance across web navigation~\cite{wang2025inducing,zhou2023webarena,chen2026cua}, computer control~\cite{kuroki2024agent}, and long-horizon planning~\cite{zabounidis2025scalar,liskilltracer}. Building upon prior passive experience reuse, recent studies have driven the dynamic injection and deep co-evolution of agentic skills through reinforcement learning and closed-loop analysis~\cite{wang2025reinforcement,xia2026skillrl,zhang2026memskill, yang2026autoskill, alzubi2026evoskill}. Concurrently, frequent skill iteration demands auditable verification to ensure lifecycle safety~\cite{jiang2026sok,huang2025audited}. As skill banks expand, researchers have introduced ontological networks, complete routing mechanisms, and advanced compositional benchmarks~\cite{liang2026skillnet, zheng2026skillrouter, chen2026skillcraft}. Despite this rapid progress, existing frameworks still treat the skill bank and the collaboration structure of multi-agent systems as largely decoupled. Skills are updated in response to task outcomes, but no signal is propagated to restructure how agents collaborate, and visual features play no role in skill retrieval or evolution. SkillGraph addresses both gaps through a co-evolution loop: the Skill Bank shapes MMGT node representations, while topology predictions feed back into skill selection signals during training, thereby making graph structure and skill knowledge mutually reinforcing for the first time.

\section{Method}
\label{sec:method}
\begin{figure}[t]
  \centering
  \includegraphics[width=\linewidth]{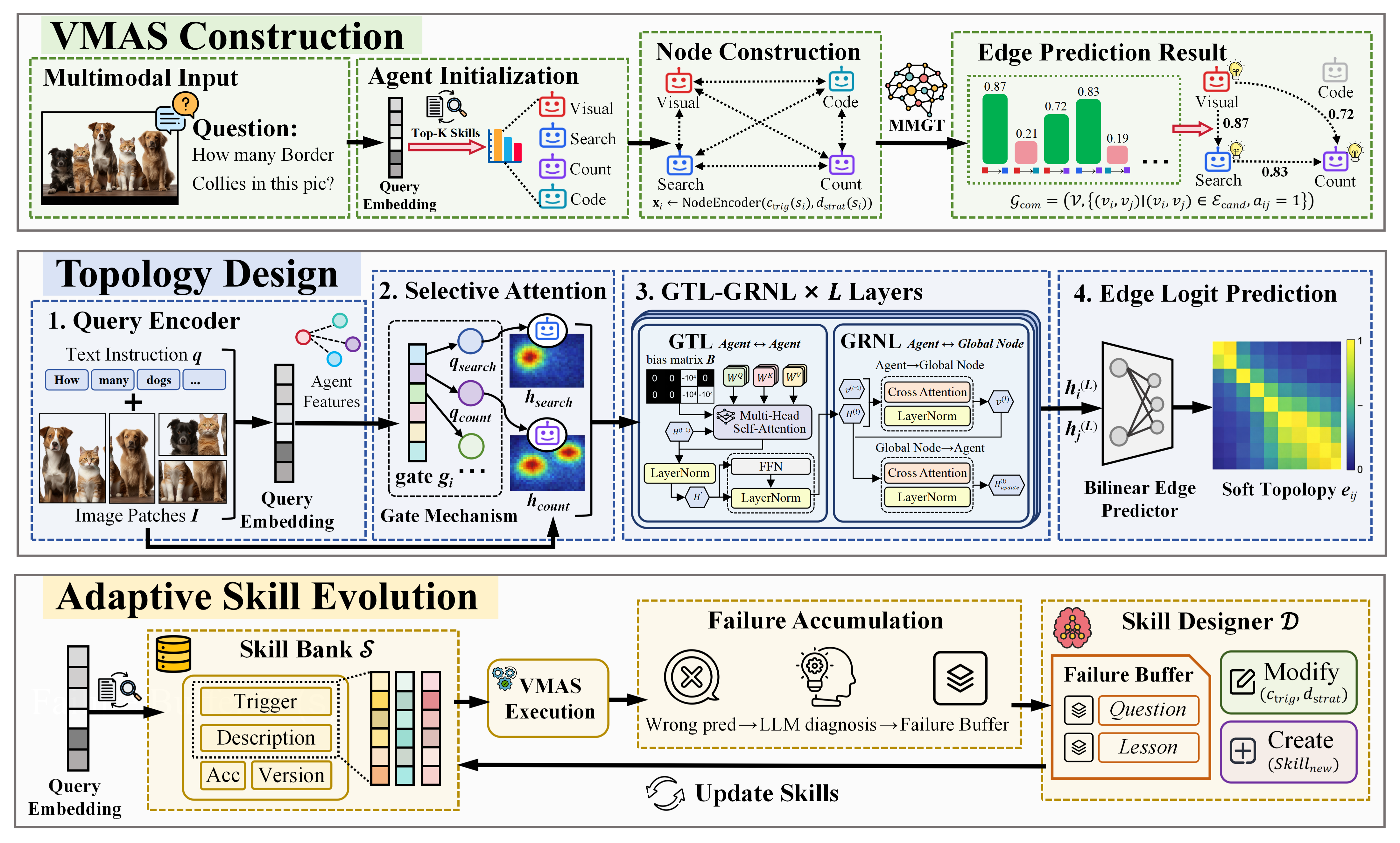}
  \caption{\textbf{SkillGraph Framework.} The system operates in three stages: \textbf{VMAS Construction:} Agents retrieve dynamic skills to initialize policy-aware node features. \textbf{Topology Design:} The Multimodal Graph Transformer (MMGT) fuses visual patches and task semantics to predict a query-conditioned communication topology. \textbf{Adaptive Skill Evolution:} A Skill Designer refines skills using failure logs. Updated skills feed directly back into the MMGT, closing the co-evolution loop.}
  \label{fig:overview}
\end{figure}

\Cref{fig:overview} illustrates the overall SkillGraph pipeline.
Given a multimodal query \(\mathcal{Q}=(q,\mathcal{I})\), the
framework operates in three coupled stages.
In the \emph{Construct} stage (\cref{sec:construct}), each agent
is equipped with a dynamically retrieved reasoning skill drawn
from a hierarchical Skill Bank \(\mathcal{S}\), and multimodal
node features are assembled to reflect each agent's instantaneous
behavioral policy.
In the \emph{Design} stage (\cref{sec:mmgt}), a Multimodal Graph
Transformer (MMGT) jointly encodes image content and question
semantics to predict a query-conditioned communication topology
\(\mathcal{G}_{\mathrm{com}}\).
In the \emph{Evolve} stage (\cref{sec:evolve}), a Skill Designer
continuously synthesizes new skills from accumulated failure
experience; the resulting updates propagate back into MMGT's node
representations, closing a co-evolution loop between agent
capability and communication structure.
The full training procedure is described in \cref{sec:training}.

\subsection{VMAS Construction}
\label{sec:construct}

\paragraph{Agent initialization.}
Let \(\mathcal{V}=\{v_i\}_{i=1}^{N}\) denote the set of
specialized agents.
Prior work assigns agents static role descriptions that remain
fixed throughout training, preventing the system from adapting
to novel visual sub-tasks that emerge at test time.
To address this, each agent \(v_i\) is instead equipped with a
reasoning \emph{skill} \(s_i \in \mathcal{S}\), dynamically
selected from the Skill Bank by semantic retrieval.
A skill is a structured tuple
\(s = (c_{\mathrm{trig}},\, d_{\mathrm{strat}},\,
\pi,\, \mathcal{F},\, \nu)\),
where \(c_{\mathrm{trig}}\) describes the visual sub-task for which
the skill is suited, \(d_{\mathrm{strat}}\) provides step-by-step
reasoning instructions, \(\pi = n_{\mathrm{succ}}/n_{\mathrm{use}}\)
is a running accuracy estimate, \(\mathcal{F}\) is a bounded failure
buffer, and \(\nu\) is a version counter used to
track evolution history.
The failure buffer \(\mathcal{F}(s)\) stores structured failure records
\(\bigl(q,\,\mathcal{I},\,\hat{a},\,a^*,\,l\bigr)\) with bounded capacity.
This structured representation enables the Skill Designer
(\cref{sec:evolve}) to perform evidence-based revisions
rather than unconstrained free-form rewrites.

\paragraph{Node feature construction.}
Since node features must reflect each agent's current reasoning
strategy, we encode the active skill
of each role using a lightweight sentence
encoder:
\begin{equation}
  \mathbf{x}_i \;\leftarrow\;
  \mathrm{NodeEncoder}\!\bigl(c_{\mathrm{trig}}(s_i),\,
  d_{\mathrm{strat}}(s_i)\bigr) \;\in\; \mathbb{R}^{D},
  \label{eq:node_feat}
\end{equation}
where the concatenation of the skill's trigger condition and the strategy text serves as the encoding
input.
This design ensures that \emph{node features track the agents'
instantaneous behavioral policies}: whenever a skill is retrieved
or evolved, \cref{eq:node_feat} is re-evaluated and the updated
embedding is propagated into the topology predictor \emph{without
any additional parameter update}, incurring negligible overhead.
The full feature matrix is
\(\mathbf{X}_{\mathrm{agent}} =
[\mathbf{x}_1,\ldots,\mathbf{x}_N]^\top \in \mathbb{R}^{N \times D}\).

\subsection{ Multimodal Graph Topology Design}
\label{sec:mmgt}

Building upon \(\mathbf{X}_{\mathrm{agent}}\) and
\(\mathbf{A}_{\mathrm{role}}\), SkillGraph must establish a
fine-grained, query-specific communication topology
\(\mathcal{G}_{\mathrm{com}}\).
For multimodal tasks, the optimal communication structure depends
critically on \emph{what is in the image}: inputs with dense text call for OCR-oriented collaboration, while complex spatial layouts demand different routing among perception and reasoning agents.
Conditioning the topology predictor on textual agent profiles
alone is therefore insufficient.
To capture this visual dependency, we introduce the Multimodal
Graph Transformer (MMGT), a five-stage encoder that jointly
processes image patches, question text, and inter-agent role
priors to produce pairwise edge logits.

\subsubsection{Multimodal Query Encoder.}
We augment the multi-agent structure with a task-specific query encoder
\(v_{\mathrm{task}}\) that fuses the question embedding with image patch features into a shared global context for all agents.
A frozen clip encoder extracts \(P\) patch tokens
\(\mathbf{Q}_{\mathrm{img}} \in \mathbb{R}^{P \times D_{\mathrm{img}}}\)
from \(\mathcal{I}\), and \(q\) is encoded as
\(\mathbf{q}_{\mathrm{text}} \in \mathbb{R}^{D}\) by the sentence
encoder.
The embedding is obtained via cross-attention with
a residual text shortcut:
\begin{equation}
  \mathbf{v} \;=\;
  \mathrm{LayerNorm}\!\Bigl(
    \mathrm{CrossAttn}\bigl(
      \mathbf{W}_{\mathrm{t}}\mathbf{q}_{\mathrm{text}},\;
      \mathbf{W}_{\mathrm{img}}\mathbf{Q}_{\mathrm{img}}
    \bigr)
    + \mathbf{W}_{\mathrm{t}}\mathbf{q}_{\mathrm{text}}
  \Bigr) \;\in\; \mathbb{R}^{d},
  \label{eq:vnode}
\end{equation}
where \(\mathbf{W}_{\mathrm{t}} \in \mathbb{R}^{d \times D}\) and
\(\mathbf{W}_{\mathrm{img}} \in \mathbb{R}^{d \times D_{\mathrm{img}}}\)
are learned projections.
The residual term \(\mathbf{W}_{\mathrm{t}}\mathbf{q}_{\mathrm{text}}\)
ensures that question semantics are never washed out by dominant
image features---a failure mode we observed when using
cross-attention alone.

\subsubsection{Per-Agent Selective Image Attention.}
\label{sec:per_agent}
Even with a multimodal virtual node, a second limitation remains:
providing all agents with the \emph{same} global image
representation prevents role-specialized visual grounding.
SkillGraph resolves this by letting each agent attend
\emph{independently} to the image regions relevant to its
assigned skill.
We first project each agent's skill embedding into the working
representation space:
\begin{equation}
  \mathbf{h}_i^{(0)} \;=\;
  \mathrm{GELU}\!\bigl(\mathrm{LayerNorm}(
    \mathbf{W}_{\mathrm{node}}\,\mathbf{x}_i)\bigr)
  \;\in\; \mathbb{R}^{d}.
  \label{eq:proj}
\end{equation}
The query embedding \(\mathbf{v}\), which encodes the global task context,
then modulates each agent's image query through a gating mechanism:
\begin{align}
  \mathbf{g}_i \;&=\;
    \sigma\!\Bigl(
      \mathbf{W}_{g}\bigl[\mathbf{h}_i^{(0)} \,\|\, \mathbf{v}\bigr]
    \Bigr),
  \label{eq:gate} \\
  \tilde{\mathbf{q}}_i \;&=\;
    \mathbf{h}_i^{(0)} + \mathbf{g}_i \odot \mathbf{v},
  \label{eq:gated_q}
\end{align}
where \(\sigma\) is the sigmoid function and \(\|\) denotes
concatenation.
The gate \(\mathbf{g}_i \in [0,1]^d\) learns how much global
task context each agent should incorporate into its local image
query: For example, OCR-related skills learns near-unity
gates for text-salient patch regions, while counting
skills activates gates for object-dense areas.
The updated node features after selective image attention are:
\begin{equation}
  \mathbf{h}_i^{(1)} \;=\;
    \mathrm{LayerNorm}\!\Bigl(
      \mathbf{h}_i^{(0)}
      + \mathrm{CrossAttn}\!\bigl(
          \tilde{\mathbf{q}}_i,\;
          \mathbf{W}_{\mathrm{img}}\mathbf{Q}_{\mathrm{img}}
        \bigr)
    \Bigr).
  \label{eq:img_attn}
\end{equation}
The residual connection in \cref{eq:img_attn} preserves the
original skill-derived features while additively enriching them
with role-relevant visual context.

\subsubsection{Graph Transformer with Role-Prior Bias.}
\label{sec:gt}
To model inter-agent dependencies, we stack \(L\) alternating
Graph Transformer Layers (GTL) and Global Relay Node Layers (GRNL).
Each GTL performs full pairwise self-attention over all agents,
with the attention scores for each pair modulated by an additive
role-prior bias before softmax:
\begin{equation}
  \mathbf{H}' \;=\; \mathrm{LayerNorm}\!\Bigl(
    \mathbf{H} + \mathrm{softmax}\!\Bigl(
      \frac{\mathbf{H}W_Q\,(\mathbf{H}W_K)^\top}{\sqrt{d}}
      + \mathbf{B}
    \Bigr)\mathbf{H}W_V\,W_O
  \Bigr),
  \label{eq:gtl_attn}
\end{equation}
\begin{equation}
  \mathbf{H}^{(\ell)} \;=\; \mathrm{LayerNorm}\!\bigl(
    \mathbf{H}' + \mathrm{FFN}(\mathbf{H}')
  \bigr),
  \label{eq:gtl_ffn}
\end{equation}
where \(\mathrm{FFN}\) uses GELU activations and pre-norm residuals,
and the bias matrix \(\mathbf{B}\) is constructed from
\(\mathbf{A}_{\mathrm{role}}\) as:
\begin{equation}
  B_{ij} \;=\; \begin{cases}
    \phantom{-}0 & (i,j) \in \mathbf{A}_{\mathrm{role}}, \\
    -10^{4}      & \text{otherwise.}
  \end{cases}
  \label{eq:bias}
\end{equation}
Rather than applying \(\mathbf{A}_{\mathrm{role}}\) as a hard
mask that permanently forbids cross-role communication,
the bias in \cref{eq:bias} merely \emph{discourages}
role-incompatible pairs: their attention logits are suppressed by
a large negative value before softmax, making cross-role attention
improbable but not impossible.
Whenever multimodal evidence is strong enough, the model can
overcome this suppression and still route information across
role boundaries, while in low-evidence settings the human
structural prior naturally dominates.

\subsubsection{Global Relay Node Bidirectional Interaction.}
After each GTL, a GRNL implements two-way information exchange
between \(\mathbf{v}\) and the full agent feature set
\(\mathbf{H}_{\mathrm{temp}}^{(\ell)} = [\mathbf{h}_1^{(\ell)}, \ldots,
\mathbf{h}_N^{(\ell)}]^\top\):
\begin{align}
  \mathbf{v}^{(\ell)} \;&=\;
    \mathrm{LayerNorm}\!\bigl(
      \mathbf{v}^{(\ell-1)} +
      \mathrm{CrossAttn}(\mathbf{v}^{(\ell-1)},\,
        \mathbf{H}^{(\ell)})
    \bigr),
  \label{eq:nodes2vn} \\[2pt]
  \mathbf{H}_{\mathrm{update}}^{(\ell)} \;&\leftarrow\;
    \mathrm{LayerNorm}\!\bigl(
      \mathbf{H}^{(\ell)} +
      \mathrm{CrossAttn}(\mathbf{H}^{(\ell)},\,
        \mathbf{v}^{(\ell)})
    \bigr).
  \label{eq:vn2nodes}
\end{align}
\Cref{eq:nodes2vn} aggregates the collective agent state into the
global relay node, while \cref{eq:vn2nodes} broadcasts the updated
global task context back to every agent.
This bidirectional design ensures that \(\mathbf{v}^{(\ell)}\)
accumulates progressively richer collaborative state across layers,
preventing it from acting as a passive information bottleneck.

\subsubsection{Edge Logit Prediction.}
After \(L\) GTL-GRNL rounds, pairwise edge logits are computed via
a direction-aware bilinear predictor:
\begin{equation}
  e_{ij} \;=\;
    (\mathbf{h}_i^{(L)})^\top
    \mathbf{W}_{\mathrm{edge}}\,
    \mathbf{h}_j^{(L)} + b,
  \label{eq:edge}
\end{equation}
where \(\mathbf{W}_{\mathrm{edge}} \in \mathbb{R}^{d \times d}\)
and \(b \in \mathbb{R}\) are learnable parameters.
Because \(\mathbf{W}_{\mathrm{edge}}\) is not constrained to be
symmetric, the predictor natively models directed communication:
the score for edge \((v_i \to v_j)\) can differ from that of
\((v_j \to v_i)\), capturing the asymmetric information flow
characteristic of specialist agent pipelines.
The full logit vector \(\mathbf{e} \in \mathbb{R}^{N^2}\) is normalized to \([-1, 1]\) to keep edge probabilities away
from saturation and stabilize policy-gradient training.

\paragraph{Communication graph sampling.}
Sampling operates over a pre-defined candidate edge set
\(\mathcal{E}_{\mathrm{cand}} \subseteq \mathcal{V} \times \mathcal{V}\),
derived from mode-specific structural priors; edges outside \(\mathcal{E}_{\mathrm{cand}}\) are
permanently suppressed and receive no gradient signal.
Within \(\mathcal{E}_{\mathrm{cand}}\), each directed edge
\((v_i, v_j)\) is independently sampled as
\(a_{ij} \sim \mathrm{Bernoulli}(\sigma(\tilde{e}_{ij}))\),
subject to a cycle-avoidance check via BFS.
Enforcing a DAG structure ensures agents execute in a well-defined
topological order, which is necessary for correct inter-agent
message passing.
The resulting \emph{communication topology} is:
\begin{equation}
  \mathcal{G}_{\mathrm{com}} \;=\;
  \bigl(\mathcal{V},\;
  \{(v_i, v_j) \mid (v_i,v_j)\in\mathcal{E}_{\mathrm{cand}},\;
  a_{ij}=1\;).
  \label{eq:gcom}
\end{equation}

\subsection{Adaptive Skill Evolution}
\label{sec:evolve}

Multimodal tasks require a diverse range of reasoning
strategies that no single static role can cover, and a skill set
initialized solely from human knowledge will inevitably encounter
visual sub-tasks it cannot handle well.
To address both limitations within a unified framework, we maintain
a Skill Bank \(\mathcal{S} = \{s_k\}_{k=1}^{|\mathcal{S}|}\) as a
persistent repository of multimodal reasoning heuristics, and couple it
with a Skill Designer \(\mathcal{D}\), a meta-agent that
continuously refines \(\mathcal{S}\) from accumulated failure
experience, together with MMGT whose topology prediction both depends on
and in turn shapes which skills are exercised.
Together, these three components form a \emph{closed co-evolution
loop}.

\subsubsection{Skill Retrieval.}
At inference time, a single semantic retrieval is performed over
the skill library and shared across all agents.
Each skill is pre-encoded as the sentence embedding of its
trigger condition \(c_{\mathrm{trig}}\) concatenated with its strategy description \(d_{\mathrm{strat}}\),
forming a matrix \(\mathbf{M} \in \mathbb{R}^{|\mathcal{S}| \times D}\).
The query \(q\) is compared against \(\mathbf{M}\) to produce a
ranked list of \(N\) candidates, which are assigned to agents
by rank index, ensuring every agent holds a distinct skill role
within each inference round.

\subsubsection{Failure Accumulation.}
After each query, the correctness outcome is attributed to each
participating agent's active skill.
For an incorrect prediction, an LLM is prompted to generate a
concise diagnostic lesson \(l\), summarising \emph{why} the skill
failed on the given question.
We store each failure as a structured record
\(\bigl(q,\,\mathcal{I},\,\hat{a},\,a^*,\,l\bigr)\) and append it to
the skill's failure buffer \(\mathcal{F}(s_i)\).
These accumulated records serve as structured error signals for
subsequent skill evolution.

\subsubsection{Skill Evolution.}
Every \(K\) training iterations, \(\mathcal{D}\) identifies
\emph{hard skills}
\(\mathcal{S}_{\mathrm{hard}} =
\{s \in \mathcal{S} \mid |\mathcal{F}(s)| \geq \tau_f\}\), ensuring that evolution is triggered only by
consistent failure patterns rather than isolated anomalies.
For each \(s \in \mathcal{S}_{\mathrm{hard}}\), \(\mathcal{D}\)
receives the skill definition and its failure buffer
\(\mathcal{F}(s)\), and produces one of two targeted actions:

\begin{itemize}
  \item \textbf{Modify.}
  If the failure pattern indicates an imprecise strategy or trigger
  condition, \(\mathcal{D}\) revises \(d_{\mathrm{strat}}\)
  and \(c_{\mathrm{trig}}\) in place, increments \(\nu\), and resets
\(\mathcal{F}(s) \leftarrow \emptyset\) to collect fresh diagnostic evidence
under the improved specification, retaining the cumulative running accuracy
\(\pi\) across all versions.

  \item \textbf{Create.}
  If the failure pattern exposes a visual reasoning sub-task that no
  existing skill adequately covers,
  \(\mathcal{D}\) synthesizes a new skill \(s_{\mathrm{new}}\) and appends it to
  \(\mathcal{S}\).
  Once appended to \(\mathcal{S}\), \(s_{\mathrm{new}}\) participates
  in the standard semantic retrieval; its accuracy estimate \(\pi\)
  is initialized to zero and updated incrementally as queries are
  routed to it, allowing performance to converge before it displaces
  established skills.
\end{itemize}

\subsubsection{Co-Evolution Loop.}
\label{sec:coevolution}
The distinguishing contribution of SkillGraph is the
\emph{bidirectional coupling} between skill evolution and topology
design, in contrast to prior systems where agent capabilities and
communication structures are optimized independently.

\noindent
\(\mathcal{S} \to \mathrm{MMGT}\):
When any skill in \(\mathcal{S}\) is modified or created, the
embedding cache \(\mathbf{M}\) is invalidated and reconstructed
on the next forward pass.
Since node features (\cref{eq:node_feat}) and their projections
(\cref{eq:proj}) are derived from the current skill text,
MMGT's attention patterns---including the per-agent image queries
in \cref{eq:gated_q} and the role-prior bias in \cref{eq:bias}---
are implicitly updated without any parameter gradient step.
Skill evolution thus continuously reshapes the topology prediction
manifold, adapting communication structures to the enriched agent
capability set.

\noindent
\(\mathrm{MMGT} \to \mathcal{S}\):
The communication topology \(\mathcal{G}_{\mathrm{com}}\)
determined by MMGT governs which agents collaborate on each query,
directly shaping which skills are exercised on which visual
sub-tasks, and hence which failure records accumulate in
\(\mathcal{F}(\cdot)\).
A better topology channels queries to more suitable agents,
yielding more informative failure attribution and faster skill
improvement.

This mutual reinforcement---to our knowledge the first such
mechanism in multimodal fine-tuning beyond the initial seed skills.

\subsection{Training Objective}
\label{sec:training}

Since the graph sampling step in \cref{eq:gcom} is discrete and
non-differentiable, direct gradient descent through
\(\mathcal{G}_{\mathrm{com}}\) is intractable.
We therefore optimize MMGT with treating edge sampling as
a stochastic policy.
The training objective is:
\begin{equation}
  \arg\max_{\Theta}\;
  \mathbb{E}_{\Theta}\!\left[
    u\!\left(\mathcal{G}_{\mathrm{com}}(\mathcal{Q})\right)
  \right],
  \label{eq:obj}
\end{equation}
where \(\Theta\) denotes all MMGT parameters and
\(u(\cdot) \in \{0,1\}\) is the binary correctness utility.
We approximate the gradient of \cref{eq:obj} over a mini-batch
of \(B\) queries as:
\begin{equation}
  \nabla_\Theta\,\mathcal{L}_{\mathrm{MMGT}} \;\approx\;
  -\frac{1}{B}\sum_{b=1}^{B} r_b \cdot
  \nabla_\Theta \log P(\mathcal{G}_{\mathrm{com}}^{(b)}),
  \label{eq:pg}
\end{equation}
where \(r_b = u(\hat{a}_b)\) is treated as a reward scalar
(\texttt{stop\_grad}) and
\begin{equation}
  \log P\!\left(\mathcal{G}_{\mathrm{com}}^{(b)}\right) \;=\;
  \sum_{i,j}
  \Bigl[
    a_{ij}^{(b)}\log\sigma(\tilde{e}_{ij}^{(b)})
    + \bigl(1-a_{ij}^{(b)}\bigr)
      \log\bigl(1-\sigma(\tilde{e}_{ij}^{(b)})\bigr)
  \Bigr].
  \label{eq:logp}
\end{equation}
\Cref{eq:pg,eq:logp} work together to implement the algorithm update step:
topologies that yield correct answers increase the probability of
the sampled edges, while incorrect topologies decrease it in
proportion to edge probabilities.
\(r_b\) is detached from the computation graph, only
the log-probability term in \cref{eq:logp} produces gradients
back through \(\mathbf{e}^{(b)}\) to \(\Theta\). The complete procedure is summarized in \cref{alg:training}.

\begin{algorithm}[tb]
\caption{SkillGraph Training}
\label{alg:training}
\small
\begin{algorithmic}[1]
  \Require Graph $\mathcal{G}$, Skill Bank $\mathcal{S}$,
           Skill Designer $\mathcal{D}$, dataset $\mathcal{T}$,
           Adam$(\Theta, \eta)$, total iterations $T$, batch size $B$,
           evolve period $K$
  \Ensure Optimized MMGT parameters $\Theta$; evolved Skill Bank $\mathcal{S}$
  \State Initialize $\mathcal{G}$, $\mathcal{S}$, embedding cache $\mathbf{M}$
  \For{$t = 1, \ldots, T$}
    \State Sample mini-batch
           $\{(q_b, \mathcal{I}_b, a^*_b)\}_{b=1}^{B}$ from $\mathcal{T}$
    \State \textbf{-- Parallel inference (async) --}
    \For{$b = 1, \ldots, B$}
      \State $\hat{\mathcal{G}}^{(b)} \leftarrow \mathrm{deepcopy}(\mathcal{G})$;
             $\;\hat{\mathcal{G}}^{(b)}.\Theta \leftarrow \mathcal{G}.\Theta$
             \quad\textit{// shared params}
      \State Retrieve top-\(N\) skills from \(\mathbf{M}\) by semantic
           similarity; assign to agents by rank;
           rebuild \(\mathbf{X}_{\mathrm{agent}}\) via \cref{eq:node_feat}
      \State $\mathbf{e}^{(b)} \leftarrow
             \mathrm{MMGT}(\mathbf{X}_{\mathrm{agent}},
             \mathbf{A}_{\mathrm{role}},
             \mathbf{q}_{\mathrm{text}}, \mathbf{Q}_{\mathrm{img}})$
      \State Sample $\mathcal{G}_{\mathrm{com}}^{(b)}$ via \cref{eq:gcom};
             execute agents in topological order
      \State Obtain $\hat{a}_b$;\quad
             $r_b \leftarrow \mathbf{1}[\hat{a}_b = a^*_b]$
      \State Record failure record $\bigl(q_b, \mathcal{I}_b, \hat{a}_b, a^*_b, l_b\bigr)$ in $\mathcal{F}(s_i)$
             for each agent $v_i$
    \EndFor
    \State Compute $\mathcal{L}_{\mathrm{MMGT}}$ via \cref{eq:pg,eq:logp}
    \State $\Theta \leftarrow \Theta
           - \eta\,\nabla_\Theta\,\mathcal{L}_{\mathrm{MMGT}}$
    \State \textbf{-- Skill evolution (async, every $K$ steps) --}
    \If{$t \bmod K = 0$}
      \State $\mathcal{S} \leftarrow \mathcal{D}.\mathrm{evolve}(\mathcal{S})$
             \quad\textit{// modify or create skills}
      \State Invalidate embedding cache $\mathbf{M}$;
             schedule rebuild on next query
    \EndIf
  \EndFor
\end{algorithmic}
\end{algorithm}

\section{Experiments}
\subsection{Settings}

\paragraph{Baselines.}
To demonstrate the superiority of SkillGraph, we compare it against both single- and multi-agent approaches. The single-agent baseline performs standard autoregressive decoding. For the multi-agent baselines, we evaluate systems utilizing conventional inter-agent text-based information transmission across five widely adopted topology structures: Linear, Layered, Centralized, Random, and Complete. Our primary evaluations are conducted using Qwen3-VL-8B \cite{bai2025qwen3}. To further validate the structural generalizability, we extend our evaluation to diverse state-of-the-art vision-language backbones, including LLaVA-OneVision-Qwen2-7B~\cite{li2024llava}, Qwen2.5-VL-7B-Instruct~\cite{bai2025qwen25vltechnicalreport}, and InternVL3-8B~\cite{zhu2025internvl3}.

\paragraph{Benchmarks.}
Our evaluations cover four comprehensive multimodal benchmarks, including MMBench\cite{liu2024mmbench}, MathVista\cite{lu2023mathvista}, RealWorldQA, and InfoVQA\cite{mathew2022infographicvqa}, assessing perception, reasoning, and knowledge-intensive understanding. We report Accuracy (Acc.) on MMBench, MathVista, and RealWorldQA, and use Average Normalized Levenshtein Similarity(ANLS) on InfoVQA, which measures soft string-matching between predicted and ground-truth answers and is standard for text-centric VQA with OCR-style outputs.

\subsection{Main Results}

\begin{table}[t]
\centering
\caption{Performance of multi-agent graph topology structures on Qwen3-VL-8B-Instruct/Thinking. Each topology baseline is compared with our proposed SkillGraph enhancement.}
\label{tab:multiagent}
\renewcommand{\arraystretch}{1.1}
\resizebox{\textwidth}{!}{%
\begin{tabular}{l cc cc cc cc cc}
\toprule
\multirow{2}{*}{\textbf{Method}}
  & \multicolumn{2}{c}{\textbf{MMBench}}
  & \multicolumn{2}{c}{\textbf{MathVista}}
  & \multicolumn{2}{c}{\textbf{RealWorldQA}}
  & \multicolumn{2}{c}{\textbf{InfoVQA}}
  & \multicolumn{2}{c}{\textbf{Average}} \\
\cmidrule(lr){2-3}\cmidrule(lr){4-5}\cmidrule(lr){6-7}\cmidrule(lr){8-9}\cmidrule(lr){10-11}
 & Instruct & Thinking & Instruct & Thinking
 & Instruct & Thinking & Instruct & Thinking 
 & Instruct & Thinking \\
\midrule

DirectAnswer  
& 84.2 \phantom{\up{+0.0}} & 85.1 \phantom{\up{+0.0}} & 77.3 \phantom{\up{+0.0}} & 81.2 \phantom{\up{+0.0}} & 71.4 \phantom{\up{+0.0}} & 73.7 \phantom{\up{+0.0}} & 83.2 \phantom{\up{+0.0}} & 84.8 \phantom{\up{+0.0}} & 79.0 \phantom{\up{+0.0}} & 81.2 \phantom{\up{+0.0}} \\
\midrule

Linear        
& 83.7 \phantom{\up{+0.0}} & 83.0 \phantom{\up{+0.0}} & 79.7 \phantom{\up{+0.0}} & 81.7 \phantom{\up{+0.0}} & 73.4 \phantom{\up{+0.0}} & 73.2 \phantom{\up{+0.0}} & 82.7 \phantom{\up{+0.0}} & 84.7 \phantom{\up{+0.0}} & 79.9 \phantom{\up{+0.0}} & 80.7 \phantom{\up{+0.0}} \\
\rowcolor{gray!15} \quad + SkillGraph 
& \textbf{85.8} \up{+2.1} & \textbf{85.9} \up{+2.9} 
& \textbf{80.8} \up{+1.1} & \textbf{83.1} \up{+1.4} 
& \textbf{74.4} \up{+1.0} & \textbf{73.8} \up{+0.6} 
& \textbf{84.2} \up{+1.5} & \textbf{85.5} \up{+0.8} 
& \textbf{81.3} \up{+1.4} & \textbf{82.1} \up{+1.4} \\
\midrule

Layered       
& 84.6 \phantom{\up{+0.0}} & 84.9 \phantom{\up{+0.0}} & 80.2 \phantom{\up{+0.0}} & 82.1 \phantom{\up{+0.0}} & 73.9 \phantom{\up{+0.0}} & 73.6 \phantom{\up{+0.0}} & 82.9 \phantom{\up{+0.0}} & 85.3 \phantom{\up{+0.0}} & 80.4 \phantom{\up{+0.0}} & 81.5 \phantom{\up{+0.0}} \\
\rowcolor{gray!15} \quad + SkillGraph 
& \textbf{86.3} \up{+1.7} & \textbf{86.4} \up{+1.5} 
& \textbf{81.5} \up{+1.3} & \textbf{83.8} \up{+1.7} 
& \textbf{75.4} \up{+1.5} & \textbf{75.5} \up{+1.9} 
& \textbf{84.5} \up{+1.6} & \textbf{86.2} \up{+0.9} 
& \textbf{81.9} \up{+1.5} & \textbf{83.0} \up{+1.5} \\
\midrule

Centralized   
& 84.3 \phantom{\up{+0.0}} & 84.1 \phantom{\up{+0.0}} & 80.5 \phantom{\up{+0.0}} & 81.6 \phantom{\up{+0.0}} & 73.2 \phantom{\up{+0.0}} & 73.3 \phantom{\up{+0.0}} & 82.8 \phantom{\up{+0.0}} & 84.9 \phantom{\up{+0.0}} & 80.2 \phantom{\up{+0.0}} & 81.0 \phantom{\up{+0.0}} \\
\rowcolor{gray!15} \quad + SkillGraph 
& \textbf{86.1} \up{+1.8} & \textbf{86.3} \up{+2.2} 
& \textbf{82.2} \up{+1.7} & \textbf{83.7} \up{+2.1} 
& \textbf{74.9} \up{+1.7} & \textbf{75.6} \up{+2.3} 
& \textbf{84.1} \up{+1.3} & \textbf{86.1} \up{+1.2} 
& \textbf{81.8} \up{+1.6} & \textbf{82.9} \up{+2.0} \\
\midrule

Random        
& 85.2 \phantom{\up{+0.0}} & 84.8 \phantom{\up{+0.0}} & 81.6 \phantom{\up{+0.0}} & 84.0 \phantom{\up{+0.0}} & 75.3 \phantom{\up{+0.0}} & 75.9 \phantom{\up{+0.0}} & 84.9 \phantom{\up{+0.0}} & 86.0 \phantom{\up{+0.0}} & 81.8 \phantom{\up{+0.0}} & 82.7 \phantom{\up{+0.0}} \\
\rowcolor{gray!15} \quad + SkillGraph 
& \textbf{86.6} \up{+1.4} & \textbf{87.2} \up{+2.4} 
& \textbf{82.8} \up{+1.2} & \textbf{85.2} \up{+1.2} 
& \textbf{76.1} \up{+0.8} & \textbf{76.9} \up{+1.0} 
& \textbf{85.4} \up{+0.5} & \textbf{86.4} \up{+0.4} 
& \textbf{82.7} \up{+1.0} & \textbf{83.9} \up{+1.3} \\
\midrule

Complete      
& 84.9 \phantom{\up{+0.0}} & 84.2 \phantom{\up{+0.0}} & 81.8 \phantom{\up{+0.0}} & 83.8 \phantom{\up{+0.0}} & 75.2 \phantom{\up{+0.0}} & 75.4 \phantom{\up{+0.0}} & 85.0 \phantom{\up{+0.0}} & 85.8 \phantom{\up{+0.0}} & 81.7 \phantom{\up{+0.0}} & 82.3 \phantom{\up{+0.0}} \\
\rowcolor{gray!15} \quad + SkillGraph 
& \textbf{86.8} \up{+1.9} & \textbf{87.4} \up{+3.2} 
& \textbf{83.2} \up{+1.4} & \textbf{85.6} \up{+1.8} 
& \textbf{76.4} \up{+1.2} & \textbf{78.1} \up{+2.7} 
& \textbf{85.3} \up{+0.3} & \textbf{87.2} \up{+1.4} 
& \textbf{82.9} \up{+1.2} & \textbf{84.6} \up{+2.3} \\

\bottomrule
\end{tabular}%
}
\end{table}

\begin{figure}[t]
  \centering
  \includegraphics[width=\linewidth, height=4.1cm, keepaspectratio=false]{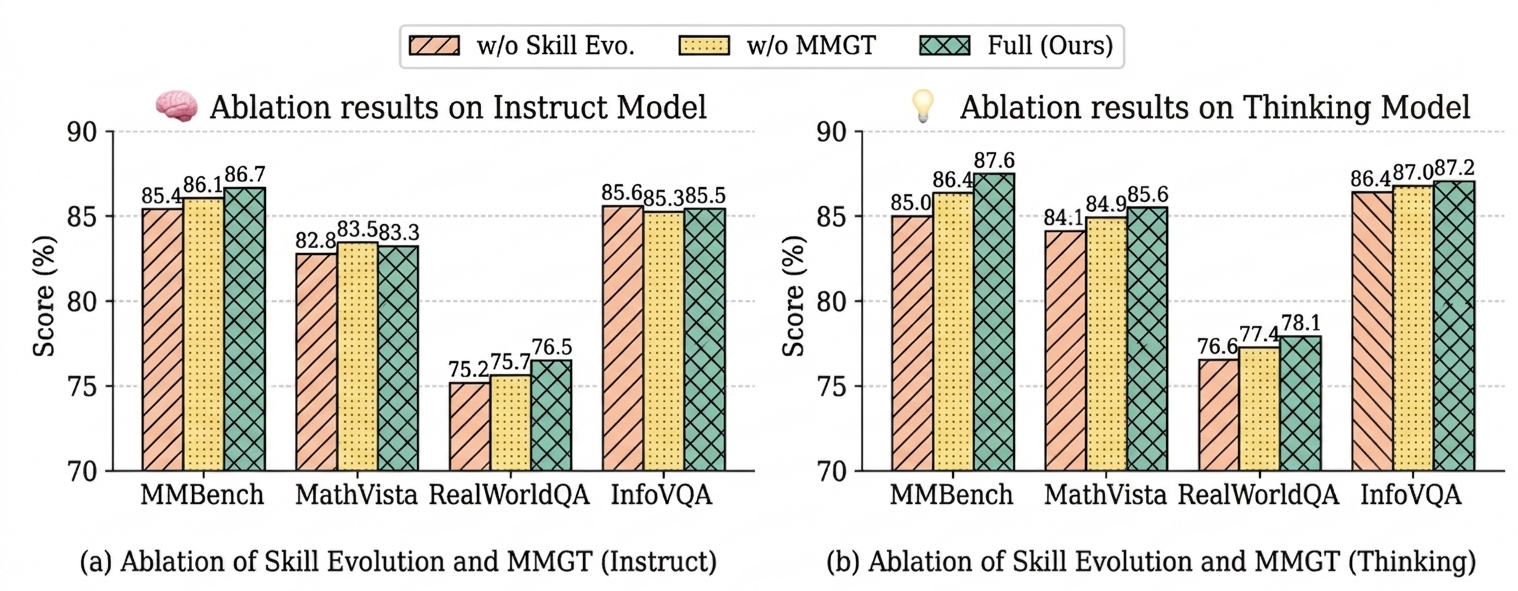}
  \caption{\textbf{Ablation study of SkillGraph components.} Evaluating the effects of \textit{Skill Evolution} and \textit{MMGT} under Instruct and Thinking settings.}
  \label{fig:ablation_instruct}
\end{figure}

\paragraph{Performance Improvements to VMAS.}
As presented in Table~\ref{tab:multiagent}, SkillGraph delivers consistent accuracy improvements across all four primary benchmarks. The gains are especially pronounced on benchmarks that place stronger demands on compositional reasoning and fine-grained visual grounding, particularly MathVista and RealWorldQA. Specifically, MathVista emphasizes mathematical reasoning in visual contexts, while RealWorldQA focuses on practical scene understanding in real-world environments, notably spatial relations and physical reasoning grounded in authentic images. MMBench, serving as a broad multimodal benchmark covering diverse perception and reasoning skills, also shows consistent improvements. By contrast, InfoVQA exhibits relatively smaller gains. We hypothesize that this is because InfoVQA is more heavily constrained by document perception, text grounding, and layout understanding, leaving comparatively less residual room for gains from topology-conditioned inter-agent collaboration than tasks requiring more diverse multi-step reasoning. Furthermore, as shown in Table~\ref{tab:model_compare}, SkillGraph consistently improves performance across diverse vision-language backbones, including LLaVA-OneVision and InternVL3, underscoring the generality and robustness of our co-evolutionary design across architectural families.

\begin{table*}[!t] 
\centering
\renewcommand{\arraystretch}{0.85} 
\caption{Performance comparison of various models and methods across multiple benchmarks. Improvements brought by the SkillGraph method are highlighted.}
\label{tab:model_compare}
\resizebox{0.9\linewidth}{!}{
\begin{tabular}{ll | c | c | c | c | c}
\toprule
\textbf{Model} & \textbf{Method} & \textbf{MMBench} & \textbf{MathVista} & \textbf{RealWorldQA} & \textbf{InfoVQA} & \textbf{Average} \\
\midrule

& DirectAnswer & 84.2\phantom{ $\uparrow$0.0} & 77.3\phantom{ $\uparrow$0.0} & 71.4\phantom{ $\uparrow$0.0} & 83.2\phantom{ $\uparrow$0.0} & 79.0\phantom{ $\uparrow$0.0} \\
& Linear & 83.7\phantom{ $\uparrow$0.0} & 79.7\phantom{ $\uparrow$0.0} & 73.4\phantom{ $\uparrow$0.0} & 82.7\phantom{ $\uparrow$0.0} & 79.9\phantom{ $\uparrow$0.0} \\
\rowcolor{gray!15} \cellcolor{white} & \quad + SkillGraph & \textbf{85.8} \textcolor{teal}{$\uparrow$2.1} & \textbf{80.8} \textcolor{teal}{$\uparrow$1.1} & \textbf{74.4} \textcolor{teal}{$\uparrow$1.0} & \textbf{84.2} \textcolor{teal}{$\uparrow$1.5} & \textbf{81.3} \textcolor{teal}{$\uparrow$1.4} \\
& Layered & 84.6\phantom{ $\uparrow$0.0} & 80.2\phantom{ $\uparrow$0.0} & 73.9\phantom{ $\uparrow$0.0} & 82.9\phantom{ $\uparrow$0.0} & 80.4\phantom{ $\uparrow$0.0} \\
\rowcolor{gray!15} \cellcolor{white} & \quad + SkillGraph & \textbf{86.3} \textcolor{teal}{$\uparrow$1.7} & \textbf{81.5} \textcolor{teal}{$\uparrow$1.3} & \textbf{75.4} \textcolor{teal}{$\uparrow$1.5} & \textbf{84.5} \textcolor{teal}{$\uparrow$1.6} & \textbf{81.9} \textcolor{teal}{$\uparrow$1.5} \\
& Centralized & 84.3\phantom{ $\uparrow$0.0} & 80.5\phantom{ $\uparrow$0.0} & 73.2\phantom{ $\uparrow$0.0} & 82.8\phantom{ $\uparrow$0.0} & 80.2\phantom{ $\uparrow$0.0} \\
\rowcolor{gray!15} \cellcolor{white} & \quad + SkillGraph & \textbf{86.1} \textcolor{teal}{$\uparrow$1.8} & \textbf{82.2} \textcolor{teal}{$\uparrow$1.7} & \textbf{74.9} \textcolor{teal}{$\uparrow$1.7} & \textbf{84.1} \textcolor{teal}{$\uparrow$1.3} & \textbf{81.8} \textcolor{teal}{$\uparrow$1.6} \\
& Random & 85.2\phantom{ $\uparrow$0.0} & 81.6\phantom{ $\uparrow$0.0} & 75.3\phantom{ $\uparrow$0.0} & 84.9\phantom{ $\uparrow$0.0} & 81.8\phantom{ $\uparrow$0.0} \\
\rowcolor{gray!15} \cellcolor{white} & \quad + SkillGraph & \textbf{86.6} \textcolor{teal}{$\uparrow$1.4} & \textbf{82.8} \textcolor{teal}{$\uparrow$1.2} & \textbf{76.1} \textcolor{teal}{$\uparrow$0.8} & \textbf{85.4} \textcolor{teal}{$\uparrow$0.5} & \textbf{82.7} \textcolor{teal}{$\uparrow$1.0} \\
& Complete & 84.9\phantom{ $\uparrow$0.0} & 81.8\phantom{ $\uparrow$0.0} & 75.2\phantom{ $\uparrow$0.0} & 85.0\phantom{ $\uparrow$0.0} & 81.7\phantom{ $\uparrow$0.0} \\
\multirow{-11}{*}{\makecell[l]{Qwen3-VL-\\8B-Instruct}} & \cellcolor{gray!15} \quad + SkillGraph & \cellcolor{gray!15} \textbf{86.8} \textcolor{teal}{$\uparrow$1.9} & \cellcolor{gray!15} \textbf{83.2} \textcolor{teal}{$\uparrow$1.4} & \cellcolor{gray!15} \textbf{76.4} \textcolor{teal}{$\uparrow$1.2} & \cellcolor{gray!15} \textbf{85.3} \textcolor{teal}{$\uparrow$0.3} & \cellcolor{gray!15} \textbf{82.9} \textcolor{teal}{$\uparrow$1.2} \\
\midrule

& DirectAnswer & 81.6\phantom{ $\uparrow$0.0} & 64.8\phantom{ $\uparrow$0.0} & 67.2\phantom{ $\uparrow$0.0} & 71.3\phantom{ $\uparrow$0.0} & 71.2\phantom{ $\uparrow$0.0} \\
& Linear & 81.4\phantom{ $\uparrow$0.0} & 66.5\phantom{ $\uparrow$0.0} & 68.1\phantom{ $\uparrow$0.0} & 71.1\phantom{ $\uparrow$0.0} & 71.8\phantom{ $\uparrow$0.0} \\
\rowcolor{gray!15} \cellcolor{white} & \quad + SkillGraph & \textbf{83.2} \textcolor{teal}{$\uparrow$1.8} & \textbf{68.7} \textcolor{teal}{$\uparrow$2.2} & \textbf{69.6} \textcolor{teal}{$\uparrow$1.5} & \textbf{72.8} \textcolor{teal}{$\uparrow$1.7} & \textbf{73.6} \textcolor{teal}{$\uparrow$1.8} \\
& Layered & 82.1\phantom{ $\uparrow$0.0} & 67.2\phantom{ $\uparrow$0.0} & 68.5\phantom{ $\uparrow$0.0} & 71.5\phantom{ $\uparrow$0.0} & 72.3\phantom{ $\uparrow$0.0} \\
\rowcolor{gray!15} \cellcolor{white} & \quad + SkillGraph & \textbf{83.5} \textcolor{teal}{$\uparrow$1.4} & \textbf{69.2} \textcolor{teal}{$\uparrow$2.0} & \textbf{70.1} \textcolor{teal}{$\uparrow$1.6} & \textbf{72.7} \textcolor{teal}{$\uparrow$1.2} & \textbf{73.9} \textcolor{teal}{$\uparrow$1.6} \\
& Centralized & 81.7\phantom{ $\uparrow$0.0} & 67.0\phantom{ $\uparrow$0.0} & 67.9\phantom{ $\uparrow$0.0} & 71.3\phantom{ $\uparrow$0.0} & 72.0\phantom{ $\uparrow$0.0} \\
\rowcolor{gray!15} \cellcolor{white} & \quad + SkillGraph & \textbf{83.6} \textcolor{teal}{$\uparrow$1.9} & \textbf{69.5} \textcolor{teal}{$\uparrow$2.5} & \textbf{69.8} \textcolor{teal}{$\uparrow$1.9} & \textbf{72.6} \textcolor{teal}{$\uparrow$1.3} & \textbf{73.9} \textcolor{teal}{$\uparrow$1.9} \\
& Random & 82.5\phantom{ $\uparrow$0.0} & 68.1\phantom{ $\uparrow$0.0} & 69.4\phantom{ $\uparrow$0.0} & 73.0\phantom{ $\uparrow$0.0} & 73.3\phantom{ $\uparrow$0.0} \\
\rowcolor{gray!15} \cellcolor{white} & \quad + SkillGraph & \textbf{83.8} \textcolor{teal}{$\uparrow$1.3} & \textbf{70.2} \textcolor{teal}{$\uparrow$2.1} & \textbf{70.3} \textcolor{teal}{$\uparrow$0.9} & \textbf{73.6} \textcolor{teal}{$\uparrow$0.6} & \textbf{74.5} \textcolor{teal}{$\uparrow$1.2} \\
& Complete & 82.4\phantom{ $\uparrow$0.0} & 68.4\phantom{ $\uparrow$0.0} & 69.2\phantom{ $\uparrow$0.0} & 72.7\phantom{ $\uparrow$0.0} & 73.2\phantom{ $\uparrow$0.0} \\
\multirow{-11}{*}{\begin{tabular}[c]{@{}l@{}}LLaVA-OV-\\Qwen2-7B\end{tabular}} & \cellcolor{gray!15} \quad + SkillGraph & \cellcolor{gray!15} \textbf{84.1} \textcolor{teal}{$\uparrow$1.7} & \cellcolor{gray!15} \textbf{70.8} \textcolor{teal}{$\uparrow$2.4} & \cellcolor{gray!15} \textbf{70.5} \textcolor{teal}{$\uparrow$1.3} & \cellcolor{gray!15} \textbf{73.1} \textcolor{teal}{$\uparrow$0.4} & \cellcolor{gray!15} \textbf{74.6} \textcolor{teal}{$\uparrow$1.5} \\
\midrule

& DirectAnswer & 83.9\phantom{ $\uparrow$0.0} & 67.5\phantom{ $\uparrow$0.0} & 68.6\phantom{ $\uparrow$0.0} & 82.9\phantom{ $\uparrow$0.0} & 75.7\phantom{ $\uparrow$0.0} \\
& Linear & 83.6\phantom{ $\uparrow$0.0} & 68.9\phantom{ $\uparrow$0.0} & 69.3\phantom{ $\uparrow$0.0} & 82.5\phantom{ $\uparrow$0.0} & 76.1\phantom{ $\uparrow$0.0} \\
\rowcolor{gray!15} \cellcolor{white} & \quad + SkillGraph & \textbf{85.0} \textcolor{teal}{$\uparrow$1.4} & \textbf{70.6} \textcolor{teal}{$\uparrow$1.7} & \textbf{70.6} \textcolor{teal}{$\uparrow$1.3} & \textbf{84.1} \textcolor{teal}{$\uparrow$1.6} & \textbf{77.6} \textcolor{teal}{$\uparrow$1.5} \\
& Layered & 84.2\phantom{ $\uparrow$0.0} & 69.4\phantom{ $\uparrow$0.0} & 69.7\phantom{ $\uparrow$0.0} & 82.8\phantom{ $\uparrow$0.0} & 76.5\phantom{ $\uparrow$0.0} \\
\rowcolor{gray!15} \cellcolor{white} & \quad + SkillGraph & \textbf{85.4} \textcolor{teal}{$\uparrow$1.2} & \textbf{71.5} \textcolor{teal}{$\uparrow$2.1} & \textbf{71.1} \textcolor{teal}{$\uparrow$1.4} & \textbf{84.2} \textcolor{teal}{$\uparrow$1.4} & \textbf{78.1} \textcolor{teal}{$\uparrow$1.5} \\
& Centralized & 83.8\phantom{ $\uparrow$0.0} & 69.5\phantom{ $\uparrow$0.0} & 69.2\phantom{ $\uparrow$0.0} & 82.6\phantom{ $\uparrow$0.0} & 76.3\phantom{ $\uparrow$0.0} \\
\rowcolor{gray!15} \cellcolor{white} & \quad + SkillGraph & \textbf{85.3} \textcolor{teal}{$\uparrow$1.5} & \textbf{71.1} \textcolor{teal}{$\uparrow$1.6} & \textbf{70.9} \textcolor{teal}{$\uparrow$1.7} & \textbf{84.1} \textcolor{teal}{$\uparrow$1.5} & \textbf{77.9} \textcolor{teal}{$\uparrow$1.6} \\
& Random & 84.5\phantom{ $\uparrow$0.0} & 70.2\phantom{ $\uparrow$0.0} & 70.4\phantom{ $\uparrow$0.0} & 84.1\phantom{ $\uparrow$0.0} & 77.3\phantom{ $\uparrow$0.0} \\
\rowcolor{gray!15} \cellcolor{white} & \quad + SkillGraph & \textbf{85.8} \textcolor{teal}{$\uparrow$1.3} & \textbf{72.8} \textcolor{teal}{$\uparrow$2.6} & \textbf{72.2} \textcolor{teal}{$\uparrow$1.8} & \textbf{84.6} \textcolor{teal}{$\uparrow$0.5} & \textbf{78.9} \textcolor{teal}{$\uparrow$1.6} \\
& Complete & 84.6\phantom{ $\uparrow$0.0} & 70.4\phantom{ $\uparrow$0.0} & 70.3\phantom{ $\uparrow$0.0} & 84.2\phantom{ $\uparrow$0.0} & 77.4\phantom{ $\uparrow$0.0} \\
\multirow{-11}{*}{\begin{tabular}[c]{@{}l@{}}Qwen2.5-VL-\\7B-Instruct\end{tabular}} & \cellcolor{gray!15} \quad + SkillGraph & \cellcolor{gray!15} \textbf{85.7} \textcolor{teal}{$\uparrow$1.1} & \cellcolor{gray!15} \textbf{72.7} \textcolor{teal}{$\uparrow$2.3} & \cellcolor{gray!15} \textbf{71.9} \textcolor{teal}{$\uparrow$1.6} & \cellcolor{gray!15} \textbf{84.4} \textcolor{teal}{$\uparrow$0.2} & \cellcolor{gray!15} \textbf{78.7} \textcolor{teal}{$\uparrow$1.3} \\
\midrule

& DirectAnswer & 82.2\phantom{ $\uparrow$0.0} & 71.5\phantom{ $\uparrow$0.0} & 70.7\phantom{ $\uparrow$0.0} & 77.1\phantom{ $\uparrow$0.0} & 75.4\phantom{ $\uparrow$0.0} \\
& Linear & 81.7\phantom{ $\uparrow$0.0} & 73.4\phantom{ $\uparrow$0.0} & 71.6\phantom{ $\uparrow$0.0} & 76.6\phantom{ $\uparrow$0.0} & 75.8\phantom{ $\uparrow$0.0} \\
\rowcolor{gray!15} \cellcolor{white} & \quad + SkillGraph & \textbf{83.8} \textcolor{teal}{$\uparrow$2.1} & \textbf{75.9} \textcolor{teal}{$\uparrow$2.5} & \textbf{73.4} \textcolor{teal}{$\uparrow$1.8} & \textbf{78.6} \textcolor{teal}{$\uparrow$2.0} & \textbf{77.9} \textcolor{teal}{$\uparrow$2.1} \\
& Layered & 82.5\phantom{ $\uparrow$0.0} & 74.1\phantom{ $\uparrow$0.0} & 72.2\phantom{ $\uparrow$0.0} & 76.8\phantom{ $\uparrow$0.0} & 76.4\phantom{ $\uparrow$0.0} \\
\rowcolor{gray!15} \cellcolor{white} & \quad + SkillGraph & \textbf{84.3} \textcolor{teal}{$\uparrow$1.8} & \textbf{76.4} \textcolor{teal}{$\uparrow$2.3} & \textbf{73.7} \textcolor{teal}{$\uparrow$1.5} & \textbf{78.9} \textcolor{teal}{$\uparrow$2.1} & \textbf{78.3} \textcolor{teal}{$\uparrow$1.9} \\
& Centralized & 82.1\phantom{ $\uparrow$0.0} & 73.8\phantom{ $\uparrow$0.0} & 71.5\phantom{ $\uparrow$0.0} & 76.8\phantom{ $\uparrow$0.0} & 76.1\phantom{ $\uparrow$0.0} \\
\rowcolor{gray!15} \cellcolor{white} & \quad + SkillGraph & \textbf{84.4} \textcolor{teal}{$\uparrow$2.3} & \textbf{76.5} \textcolor{teal}{$\uparrow$2.7} & \textbf{73.4} \textcolor{teal}{$\uparrow$1.9} & \textbf{78.6} \textcolor{teal}{$\uparrow$1.8} & \textbf{78.2} \textcolor{teal}{$\uparrow$2.2} \\
& Random & 82.9\phantom{ $\uparrow$0.0} & 75.2\phantom{ $\uparrow$0.0} & 73.1\phantom{ $\uparrow$0.0} & 78.5\phantom{ $\uparrow$0.0} & 77.4\phantom{ $\uparrow$0.0} \\
\rowcolor{gray!15} \cellcolor{white} & \quad + SkillGraph & \textbf{84.6} \textcolor{teal}{$\uparrow$1.7} & \textbf{77.6} \textcolor{teal}{$\uparrow$2.4} & \textbf{74.1} \textcolor{teal}{$\uparrow$1.0} & \textbf{79.1} \textcolor{teal}{$\uparrow$0.6} & \textbf{78.9} \textcolor{teal}{$\uparrow$1.4} \\
& Complete & 82.8\phantom{ $\uparrow$0.0} & 75.3\phantom{ $\uparrow$0.0} & 72.9\phantom{ $\uparrow$0.0} & 78.7\phantom{ $\uparrow$0.0} & 77.4\phantom{ $\uparrow$0.0} \\
\multirow{-11}{*}{\begin{tabular}[c]{@{}l@{}}InternVL3-\\8B\end{tabular}} & \cellcolor{gray!15} \quad + SkillGraph & \cellcolor{gray!15} \textbf{84.9} \textcolor{teal}{$\uparrow$2.1} & \cellcolor{gray!15} \textbf{77.8} \textcolor{teal}{$\uparrow$2.5} & \cellcolor{gray!15} \textbf{74.7} \textcolor{teal}{$\uparrow$1.8} & \cellcolor{gray!15} \textbf{80.2} \textcolor{teal}{$\uparrow$1.5} & \cellcolor{gray!15} \textbf{79.4} \textcolor{teal}{$\uparrow$2.0} \\
\bottomrule
\end{tabular}
}
\end{table*}

\paragraph{Ablation of Skill Evolution and MMGT.}
We ablate \textit{Skill Evolution} and \textit{MMGT} by activating each module independently and comparing them with the full SkillGraph. Fig.~\ref{fig:ablation_instruct} shows that both modules contribute positively across the majority of benchmark-setting combinations, and their combination achieves the best overall results. Skill Evolution provides consistent gains: by continuously refining and specializing the skill bank, it strengthens agent-role specialization and cross-query knowledge reuse, yielding broad improvements even without topology learning. \textit{MMGT} provides additional benefits by learning a query-specific, vision-aware communication graph, which helps agents exchange and ground information on fine-grained spatial and quantitative cues. Importantly, the full model delivers further gains beyond either single-module variant on most benchmarks, highlighting clear complementarity: Skill Evolution improves \emph{what} capabilities agents can invoke, while MMGT improves \emph{how} these capabilities are coordinated. Although a single-module variant can occasionally outperform the full model on a particular dataset, likely due to task-specific sensitivities in routing and specialization, the full model achieves the best overall trade-off across benchmarks, especially under the Thinking paradigm, where iterative deliberation benefits most from an evolving skill pool and visually conditioned communication.

\paragraph{Effect of model scale.}
Table~\ref{tab:scale_generalization} reports the performance of SkillGraph applied to backbones ranging from 2B to 38B parameters. SkillGraph yields consistent positive gains across all benchmarks and model sizes, suggesting that the co-evolution of communication topology and skill representations is broadly beneficial regardless of backbone capacity. Nevertheless, the absolute improvement generally decreases as model scale increases, a pattern consistent with the diminishing returns observed when augmenting stronger single-agent models with collaborative frameworks, since larger models already internalize richer perceptual priors and broader reasoning skills, leaving a narrower residual gap for multi-agent coordination to close. This diminishing trend is most evident in InfoVQA: as model capacity increases, improvements in document perception and text grounding may further compress the residual room available for topology-conditioned coordination. By contrast, spatial and mathematical tasks retain greater headroom for multi-agent gains even at the 32B--38B scale, where compositional sub-task diversity continues to reward dynamic routing and skill specialization.

\begin{table}[t] 
\centering 
\caption{Performance of SkillGraph across Qwen3-VL and InternVL3 models of varying scales on complete structure. Gray rows show results after applying SkillGraph to each backbone.} 
\label{tab:scale_generalization} 
\renewcommand{\arraystretch}{1.1} 
\resizebox{0.9\linewidth}{!}{%
\begin{tabular}{l ccccc} 
\toprule 
\textbf{Model}   & \textbf{MMBench}   & \textbf{MathVista}   & \textbf{RealWorldQA}   & \textbf{InfoVQA} & \textbf{Average} \\ 
\midrule  
Qwen3-VL-4B-Instruct   & 83.9\phantom{\up{+0.0}} & 76.8\phantom{\up{+0.0}} & 72.1\phantom{\up{+0.0}} & 81.8\phantom{\up{+0.0}} & 78.7\phantom{\up{+0.0}} \\ 
\cellcolor{bestgray}\quad + SkillGraph   & \cellcolor{bestgray}\textbf{86.4\up{+2.5}}   & \cellcolor{bestgray}\textbf{79.8\up{+3.0}}   & \cellcolor{bestgray}\textbf{74.3\up{+2.2}}   & \cellcolor{bestgray}\textbf{82.5\up{+0.7}} & \cellcolor{bestgray}\textbf{80.8\up{+2.1}} \\ 
\midrule  
Qwen3-VL-8B-Instruct   & 84.9\phantom{\up{+0.0}} & 81.8\phantom{\up{+0.0}} & 75.2\phantom{\up{+0.0}} & 85.0\phantom{\up{+0.0}} & 81.7\phantom{\up{+0.0}} \\ 
\cellcolor{bestgray}\quad + SkillGraph   & \cellcolor{bestgray}\textbf{86.8\up{+1.9}}   & \cellcolor{bestgray}\textbf{83.2\up{+1.4}}   & \cellcolor{bestgray}\textbf{76.4\up{+1.2}}   & \cellcolor{bestgray}\textbf{85.3\up{+0.3}} & \cellcolor{bestgray}\textbf{82.9\up{+1.2}} \\ 
\midrule  
Qwen3-VL-32B-Instruct   & 88.6\phantom{\up{+0.0}} & 84.2\phantom{\up{+0.0}} & 80.1\phantom{\up{+0.0}} & 87.6\phantom{\up{+0.0}} & 85.1\phantom{\up{+0.0}} \\ 
\cellcolor{bestgray}\quad + SkillGraph   & \cellcolor{bestgray}\textbf{89.3\up{+0.7}}   & \cellcolor{bestgray}\textbf{85.4\up{+1.2}}   & \cellcolor{bestgray}\textbf{81.2\up{+1.1}}   & \cellcolor{bestgray}\textbf{87.8\up{+0.2}} & \cellcolor{bestgray}\textbf{85.9\up{+0.8}} \\ 
\midrule  
InternVL3-2B   & 80.2\phantom{\up{+0.0}} & 59.4\phantom{\up{+0.0}} & 65.7\phantom{\up{+0.0}} & 66.8\phantom{\up{+0.0}} & 68.0\phantom{\up{+0.0}} \\ 
\cellcolor{bestgray}\quad + SkillGraph   & \cellcolor{bestgray}\textbf{83.2\up{+3.0}}   & \cellcolor{bestgray}\textbf{62.8\up{+3.4}}   & \cellcolor{bestgray}\textbf{68.4\up{+2.7}}   & \cellcolor{bestgray}\textbf{69.6\up{+2.8}} & \cellcolor{bestgray}\textbf{71.0\up{+3.0}} \\ 
\midrule  
InternVL3-8B   & 82.8\phantom{\up{+0.0}} & 75.3\phantom{\up{+0.0}} & 72.9\phantom{\up{+0.0}} & 78.7\phantom{\up{+0.0}} & 77.4\phantom{\up{+0.0}} \\ 
\cellcolor{bestgray}\quad + SkillGraph   & \cellcolor{bestgray}\textbf{84.9\up{+2.1}}   & \cellcolor{bestgray}\textbf{77.8\up{+2.5}}   & \cellcolor{bestgray}\textbf{74.7\up{+1.8}}   & \cellcolor{bestgray}\textbf{80.2\up{+1.5}} & \cellcolor{bestgray}\textbf{79.4\up{+2.0}} \\ 
\midrule  
InternVL3-38B   & 88.9\phantom{\up{+0.0}} & 76.5\phantom{\up{+0.0}} & 77.1\phantom{\up{+0.0}} & 85.7\phantom{\up{+0.0}} & 82.1\phantom{\up{+0.0}} \\ 
\cellcolor{bestgray}\quad + SkillGraph   & \cellcolor{bestgray}\textbf{89.7\up{+0.8}}   & \cellcolor{bestgray}\textbf{77.9\up{+1.4}}   & \cellcolor{bestgray}\textbf{78.2\up{+1.1}}   & \cellcolor{bestgray}\textbf{86.2\up{+0.5}} & \cellcolor{bestgray}\textbf{83.0\up{+0.9}} \\ 
\bottomrule 
\end{tabular}%
} 
\end{table}

\begin{figure}[t]
    \centering
    \includegraphics[width=\linewidth]{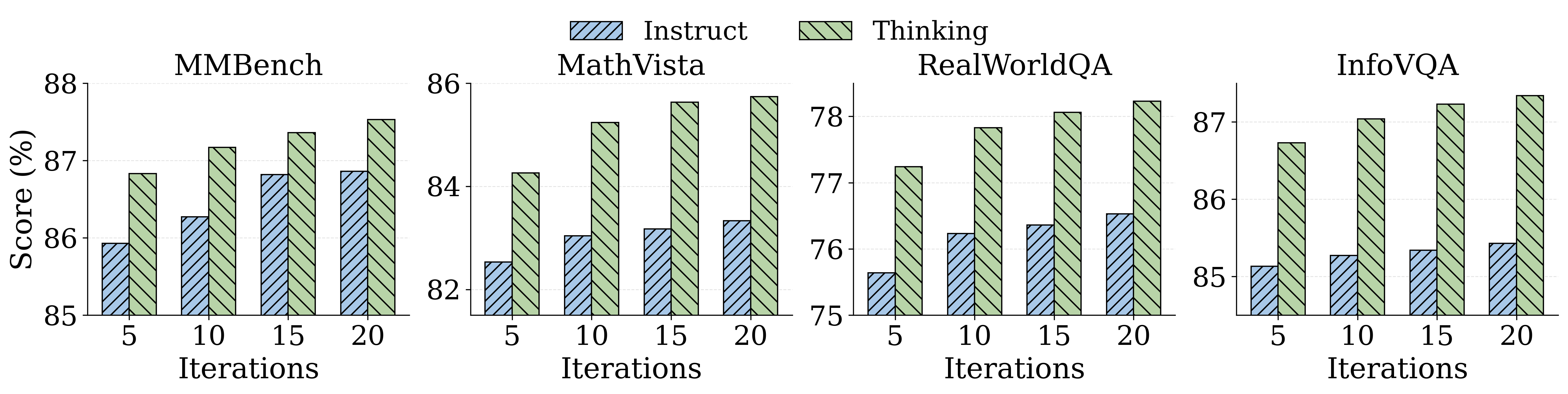}
    \caption{\textbf{Performance of SkillGraph across different iteration numbers.}}
    \label{fig:skillgraph_iterations}
\end{figure}

\paragraph{Effect of iteration number.}
Figure~\ref{fig:skillgraph_iterations} shows the performance of SkillGraph under different iteration numbers. Across all four benchmarks, performance improves consistently as the number of iterations increases, with the most noticeable gains occurring from iteration 5 to iteration 10. From iteration 10 to iteration 15, the improvements remain positive but become smaller, while the curves largely saturate after iteration 15, indicating that the most effective skill refinement and collaboration adaptation have already been acquired in the middle stage of evolution. We also observe that the Thinking setting consistently outperforms the Instruct setting on all benchmarks, suggesting that iterative deliberation benefits more from evolving skills and adaptive coordination. Among the evaluated datasets, MathVista exhibits the largest gains, especially under Thinking setting, highlighting that SkillGraph is particularly effective on reasoning-intensive multimodal tasks. RealWorldQA shows a smooth and stable upward trend, reflecting robust performance accumulation in realistic visual reasoning scenarios. By contrast, InfoVQA demonstrates smaller but still consistent improvements, implying that SkillGraph also benefits document-heavy tasks, although the overall magnitude of improvement remains more constrained by the underlying perception bottleneck. Overall, these results suggest that SkillGraph achieves substantial gains in the early and middle stages of evolution and reaches a stable performance plateau around iterations 15--20.

\begin{figure*}[t]
  \centering
  \includegraphics[width=0.9\linewidth]{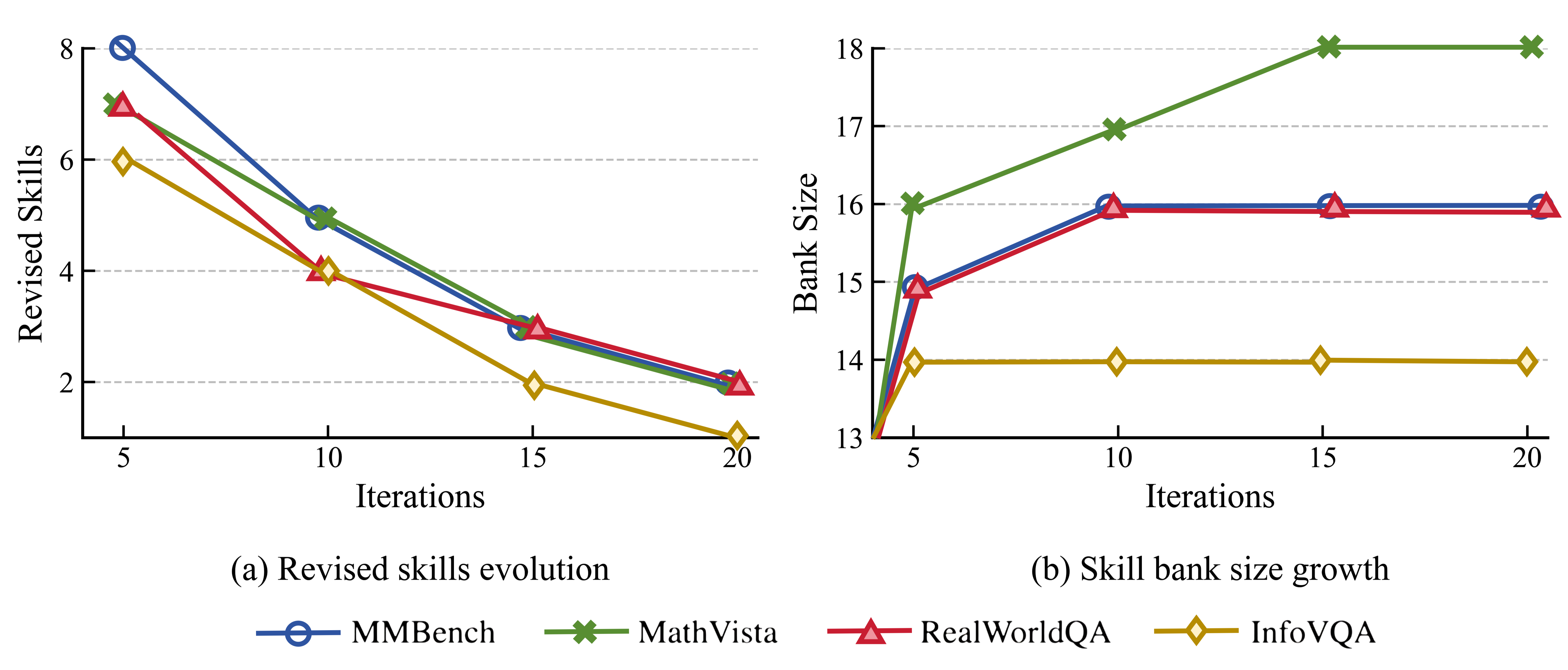}
    \caption{\textbf{Evolution of the Skill Bank across iterations.} The graph illustrates the frequency of evolution actions and the stabilization of the skill inventory across different multimodal benchmarks.}
  \label{fig:skillevo}
  \vspace{-2mm}
\end{figure*}

\begin{table*}[t]
  \centering
  \caption{Longitudinal case study of skill evolution for commonsense visual
    reasoning. 
    \colorbox{yellow!40}{\strut Yellow} highlights key iterative changes;
    \colorbox{green!25}{\strut Green} highlights diagnostic failure patterns. \(\pi\) represents the cumulative running accuracy of the skill.}
  \label{tab:app_commonsense_case}
  \renewcommand{\arraystretch}{1.15}
  \resizebox{\textwidth}{!}{%
  \begin{tabular}{p{1.8cm} p{4.2cm} p{4.8cm} p{7.2cm}}
    \toprule
    \textbf{Version}
      & \textbf{Trigger Condition \(c_{\mathrm{trig}}\)}
      & \textbf{Representative Failure \(l\)}
      & \textbf{Strategy Description \(d_{\mathrm{strat}}\)} \\
    \midrule

    Seed\newline(\(\pi=0.51\))
      & Broad questions about real-world scenes and everyday situations.
      & \textit{The skill answered ``a chef'' based on
        stereotype (white coat, kitchen setting) but the image showed a lab
        technician in a food-testing facility.}
        \textit{\hlg{Commonsense prior overrode available visual evidence.}}
      & You are strong at commonsense reasoning about real-world scenes and
        everyday situations. Use what you know about how the world works to
        interpret the image and answer the question. Ask yourself: ``What
        would normally happen in this scene?''
        \hly{Avoid letting commonsense assumptions override direct image
        evidence}, since the image may depict an unusual or staged scenario. \\

    \midrule

    v2\newline(\(\pi=0.58\))
      & Questions requiring inference of a \hly{\textit{specific attribute,
        identity, or behavior} }of a person or object from visual content.
      & \textit{Given an image of two people laughing,
        the skill inferred ``close friends'' without checking attire or
        setting cues that indicated a formal business negotiation.}
        \textit{\hlg{Inference was plausible but unsupported.}}
      & \hly{First identify what is directly visible
        and what is only inferred.} Use commonsense knowledge to generate
        plausible interpretations. \hly{Cross-check candidate answers}
        against explicit image evidence, and prefer conclusions that are
        strongly supported rather than merely typical. Be cautious with
        atypical, humorous, or staged scenes. \\

    \midrule

    v5--v7\newline(\(\pi=0.67\))
      & \hly{Questions where the answer depends on evidence-grounded inference}
        about likely attributes, roles, intentions, or actions not literally
        stated in the image, especially when multiple everyday interpretations
        are possible.
      & \textit{The skill committed to ``carrying groceries''}
        \textit{\hlg{after the first plausible hypothesis, ignoring a
        competing hypothesis}}
        \textit{(``delivering packages'') that was better
        supported by the box shape and uniform in the image.}
      & First enumerate the key visual cues
        (appearance, pose, object interaction, scene context). Then
        \hly{generate 2--3 commonsense hypotheses and test each} against the
        observed evidence. Prefer the hypothesis best supported by the
        image, not the one that is merely most stereotypical.
        \hly{If visual support is weak or ambiguous, answer conservatively}
        and avoid over-committing to fine-grained identity claims. \\

    \bottomrule
  \end{tabular}%
  }
\end{table*}

\paragraph{Analysis of Skill Bank Evolution.}
Fig.~\ref{fig:skillevo} further characterizes how the Skill Bank
evolves across benchmarks.
Two patterns stand out.
First, evolution actions (\textit{Create}\,+\,\textit{Modify}) decrease
monotonically across iterations on every benchmark, indicating that the Skill
Bank converges rather than expanding without bound.
Second, the final bank size varies by task complexity: MathVista requires the
largest inventory, reflecting the breadth of its mathematical sub-tasks
(geometry, algebra, chart reading, etc.), whereas InfoVQA saturates earliest.
This trend is consistent with the intuition that InfoVQA is an OCR-heavy
document understanding benchmark that places stronger emphasis on text grounding
and layout-aware perception than on diverse multi-step collaborative reasoning.
Once these perception and grounding capabilities are established, the
self-diagnosis loop encounters fewer novel failure modes, causing both skill
creation and modification to taper off more rapidly.

To provide a more granular perspective on how these autonomous refinements translate into performance gains, we present a qualitative case study of a specific skill's evolutionary trajectory in Table~\ref{tab:app_commonsense_case}. While Figure~\ref{fig:skillevo} demonstrates the macro-level stabilization of the skill bank, Table~\ref{tab:app_commonsense_case} illustrates the micro-level cognitive shift. As observed, the initial seed strategy heavily relies on commonsense priors, making it vulnerable to visually misleading or atypical scenarios. Prompted by self-diagnosed failure cases, the framework iteratively refines both the trigger condition ($c_{\mathrm{trig}}$) and the strategy description ($d_{\mathrm{strat}}$). By version v5-v7, the skill has evolved from making naive stereotypic assumptions to enforcing a rigorous, evidence-based hypothesis-testing protocol. This qualitative shift explains the steady quantitative improvements observed across iterations in Figure~\ref{fig:skillgraph_iterations}.

\section{Conclusion}
In this paper, we presented SkillGraph, a novel framework that overcomes the structural and cognitive bottlenecks of current VMAS by mutually optimizing agent expertise and communication topology. To replace rigid, hand-crafted routing, we introduced the Multimodal Graph Transformer (MMGT), which dynamically constructs a query-conditioned communication graph grounded in both fine-grained visual features and instruction semantics. In parallel, we proposed a Self-Evolving Skill Bank for multimodal agents, enabling the system to autonomously refine its reasoning heuristics through a continuous diagnosis of failure experiences. Crucially, SkillGraph couples these two mechanisms in a closed loop: as agent skills evolve, their updated representations can reshape the topology predictor's routing strategy without requiring additional parameter updates. Extensive experiments across diverse multimodal benchmarks and state-of-the-art vision-language backbones demonstrate that SkillGraph consistently outperforms static-topology and fixed-skill baselines, yielding particularly strong gains in complex spatial and mathematical reasoning tasks. By bridging the gap between multimodal perception, dynamic collaboration, and lifelong skill learning, SkillGraph provides a robust and scalable foundation for the next generation of collective AI systems.


\nocite{hu2025landscape,yu2025vismem,ke2025early,pan2026finscra,feng2026dr,yu2026latent,qian2026medmaslab}

\bibliographystyle{splncs04}
\bibliography{main}
\end{document}